\def\year{2019}\relax
\documentclass[letterpaper]{article} 
\usepackage{aaai19}  
\usepackage{times}  
\usepackage{helvet}  
\usepackage{courier}  
\usepackage{url}  
\usepackage{graphicx}  

\usepackage{latexsym}
\usepackage{graphicx}
\usepackage{caption}
\usepackage{xcolor}
\usepackage{colortbl}
\usepackage{hyperref} 
\usepackage{booktabs}
\usepackage{dirtytalk}
\usepackage{amsmath}
\usepackage{amssymb}
\usepackage{amsthm}
\usepackage[vlined,ruled]{algorithm2e}

\usepackage[english]{babel}

\newcolumntype{?}{!{\vrule width 2pt}}

\begin{document}
%
\title{Curriculum-Based Neighborhood Sampling For Sequence Prediction}
\author{James O' Neill and Danushka Bollegala\\
  Department of Computer Science, University of Liverpool\\
  Liverpool, L69 3BX \\
  England \\
  {\tt \{james.o-neill, danushka.bollegala\}@liverpool.ac.uk} \\}
  
\maketitle

\begin{abstract}
The task of multi-step ahead prediction in language models is challenging considering the discrepancy between training and testing. At test time, a language model is required to make predictions given past predictions as input, instead of the past targets that are provided during training. This difference, known as exposure bias, can lead to the compounding of errors along a generated sequence at test time.
\newline
In order to improve generalization in neural language models and address compounding errors, we propose a curriculum learning based method that gradually changes an initially deterministic teacher policy to a gradually more stochastic policy, which we refer to as \textit{Nearest-Neighbor Replacement Sampling}. A chosen input at a given timestep is replaced with a sampled nearest neighbor of the past target with a truncated probability proportional to the cosine similarity between the original word and its top $k$ most similar words. This allows the teacher to explore alternatives when the teacher provides a sub-optimal policy or when the initial policy is difficult for the learner to model. The proposed strategy is straightforward, online and requires little additional memory requirements. We report our main findings on two language modelling benchmarks and find that the proposed approach performs particularly well when used in conjunction with scheduled sampling, that too attempts to mitigate compounding errors in language models.

\end{abstract}

\section{Introduction}
Language modeling is a foundational challenge in natural language and bears relevance to many related tasks such as machine translation, speech recognition, image captioning and question answering ~\cite{mikolov2010recurrent,sutskever2014sequence}. It involves predicting the next token given past token/s in a sequence using a parametric model $f_{\theta}(\cdot)$ parameterized by $\theta$. This is distinctly different from standard supervised learning that assumes the input distribution to be i.i.d, which leads to a non-convex objective even when the loss is convex given the inputs.
\newline
In supervised learning the standard modeling procedure involves training a policy $\hat{\pi}$ to perform actions given an \textit{expert} policy $\pi^{*}$ at each time step $t \in T$\footnote{In the context of RL we will refer to tokens $x$ as actions $a$, models $f$ as policies $\pi$ and predictions $\hat{y}$ as decisions $\pi(s^{'}|a,s)$.}. This approach is also known as \textit{teacher forcing}~\cite{williams1989learning}. However, using the past targets $y_{t-n}$ and current input $x_t$ to predict $y_{t+1}$ at training time does not emulate the process at test time where the model is instead required to use its own past predictions $\hat{y}_{t-n}$ to generate the sequence. Sequence predictors can suffer from this discrepancy since they are not trained to use past predictions. This leads to a problem where errors compound along a generated sequence at test time, in the worst case leading to quadratic errors in $T$. Hence, learning from high dimensional and sparsely structured patterns at training time so to generate sequences from the same distribution at test time is a challenge for neural language models (NLM). Beam search is often used to somewhat address this challenge by choosing to greedily search a subset of the most probable trajectories. However, for continuous state spaces generated in NLM's this is less applicable since beam search requires the use of dynamic programming over discrete states. Additionally, this is would can be costly for large vocabularies even if used with discrete state models, particularly when used at training time and can be unsuitable for state transitions that are influenced by long-term dependencies. In such cases, learning from a teacher policy with full-supervision throughout training can be sub-optimal. 
\newline
We argue that in cases where the learner finds it difficult to learn from the teacher policy alone, sampling alternative inputs that are similar to the original input under a suitable curriculum can lead to improved out-of-sample performance and can be considered an exploration strategy. Additionally, when such a scheme is interpolated with using past predictions it can further improve performance.
We are also motivated by past findings that using outputs other than that provided from an expert can be beneficial ~\cite{hinton2015distilling,norouzi2016reward}. 
\newline
Hence, this paper proposes a curriculum learning based method whereby the teacher policy is augmented by sampling neighboring word vectors within a chosen radius with probability assigned via the normalized cosine similarities between the $k$ neighbors and a corresponding input $x_t$. By using readily available word embeddings as the basis for choosing the neighborhoods of each word, we only require to store an embedding matrix $\mathbb{R}^{|V|\times d}$ where $|V|$ is the vocabulary size with dimensionality $d$ (approx. $\log_2(|V|)$) instead of a $\mathbb{N}^{|V|\times|V|}$ transition probability matrix derived from co-occurrences. During training we monotonically increase the replacement probability using a number of simple non-parametric functions. We show that performance can be further improved when Nearest Neighbor Replacement Sampling (NNRS) is used in conjunction with scheduled sampling (SS) ~\cite{bengio2015scheduled}, a technique which also addresses the compounding errors problem. Before describing our approach in detail, we briefly introduce some of related methods in structured output prediction.

\section{Structured Output Prediction Methods}

In the past decade there has been a large effort to improve models for structured output prediction using methods from supervised learning and reinforcement learning.  

\subsection{Imitation Learning Approaches}
\textit{SEARN}~\cite{daume2009search} treats structured prediction as a search-based problem by decomposing structured prediction problems into a set of binary classifications without the need to decompose the loss function or the feature functions. A local classifier is used to sequentially (or independently) predict tokens (no actual search algorithm is involved) using past inputs and past predictions. At training time a sequence of decisions are made to perform search given an expert policy. The new policy is generated from examples provided by this policy, in which the old policy is interpolated with the new one and novel samples are assigned a cost for each token in the output. The process repeats until convergence.

\textit{Dataset Aggregation}~\cite{ross2011reduction}(DAgger) is an meta-algorithm that finds a stationary deterministic policy by allowing the model to first make predictions at test time and then asks the teacher what action they would have taken given the observed errors the model made on the observed validation data. DAgger attempts to address compounding errors by initially using an expert policy to generate a set of sequences. These samples are added to the dataset $\mathcal{D}$ and new policy is learned which subsequently generates more samples to append to $\mathcal{D}$. This process repeats until convergence and the resultant policy $\pi$ that best emulates the expert policy is used. In the initial stages of learning a modified policy is considered $\pi_i = \beta_i \pi^{*} + (1 - \beta_i)\hat{\pi}_i$ where the expert $\pi^{*}$ is used for a portion of the time (referred to as a mixture expert policy) so that the generated trajectories in the initial stages of training by $\hat{\pi}_i$ at time $i$ do not diverge to irrelevant states. Moreover, DAgger guarantees that the loss is bounded such that for a learned policy $\pi$, $J(\pi) \leq J(\pi^{*} + uTe)$ where $e$ is an upper bounded loss and $u$ is the upper bound on the increase of cost-to-go when executing a different action $a$ from either $\pi$ and $\pi^{*}$ in a $s$. This is an substantial improvement over the supervised learning guarantee, $J(\hat{\pi}) = J(\pi^{*}) + T^{2}e$ ~\cite{ross2010efficient} where the expected loss is quadratic in $T$. However, by default the re-correction steps in DAgger can be costly since it requires the dataset to grow proportional to the number of instances generated while updating the $\hat{\pi}$. 

He et al.~\shortcite{he2012imitation} propose \textit{Coaching} for imitation learning to overcome the difficulty of finding an optimal policy $\pi^{*} \in \Pi$ when there is a significant difference between the space and the experts ability, which makes it difficult to produce low error on the training data. A coach is used to learn easier actions from the expert that gradually learns more difficult actions. This can too be seen as a curriculum learning strategy. 

\textit{Aggregate Values to Imitate} (AggreVaTe)~\cite{ross2014reinforcement} extends DAgger to minimize cost-to-go of a provided expert policy instead of minimizing classification error of replicating the actions, as is the case for DAgger and Coaching. The technique is also extended to the reinforcement learning setting, providing strong guarantees for online approximate policy iteration methods. This allows a model to reason about how much cost is associated with taking a particular action in the future when subsequent actions can be from the expert policy i.e cost-to-go. In contrast, SEARN uses stochastic policies and rollouts which can be expensive. AggreVaTe instead, initially chooses a random timestep $t\in T$ from the expert policy, explores an action and observes the \textit{cost-to-go} for the expert after performing the action. AggreVaTe then trains a model to minimize expected cost from the generated cost-weighted samples. Repeating this process produces a set of cost-weighted training samples that are combined with the original dataset. The policy from the learner is then taken up to time $t$ at random, followed by an action, where the expert policy is the reintroduced from $t \to T$, again producing a new set of cost-weighted samples, and this is repeated for a chosen number of epochs.

\textit{Locally Optimal Learning To Search} (LOLS) ~\cite{chang2015learning} also aims to improve upon sub-optimal policies, while providing a local optimality guarantee in relational to the \textit{regret} on the expert policy by competing with one-step deviations from the currently learned policy $\hat{\pi}$. Two phases are taken during training: a roll-in phase that determines the distribution of states used for training and roll-out phase that affects how actions are scored. LOLS carries this out by minimizing a combination of regret to the reference policy $\pi^{r}$ and regret to one-step deviations $\pi^{d}$ represented by the second component of \autoref{eq:lols} where the regret is measured as the difference in loss $L$ between the learned policy $\hat{pi}$ and reference policy $\pi^{r}$ (i.e teacher policy) and $\pi^{d}$ respectively. 

\begin{equation}\label{eq:lols}
	\frac{1}{2}\big(L(\hat{\pi}) - L(\pi^{r})\big) + \frac{1}{2}\big(L(\hat{\pi}) - L(\pi^{d}) \big)
\end{equation}

~\cite{lampouras2016imitation} applied LOLS to generate text from unaligned data, incorporating a task-specific loss (non-differentiable) such as BLEU and ROUGE scores.

\subsection{Reinforcement Learning Approaches}

The popular policy gradient algorithm REINFORCE ~\cite{williams1992simple} has been adapted for sequence-to-sequence models by Ranzato et al. ~\shortcite{ranzato2015sequence} to allow for learning from task-specific scores (BLEU and ROUGE). The initial policy of REINFORCE is changed to allow for high dimensional action space (as is the case in our work), by instead starting at an optimal policy and slowly diverging to allow for some exploration by using the models own predictions. In fact, this strategy is similar in nature to our proposed method, particularly when using an exponential sampling function where the majority of the nearest neighbor replacements is carried out late in training and more exploration is carried out gradually throughout training.

The \textit{Actor-Critic} (AC) model ~\cite{bahdanau2016actor} proposes a critic network that predicts the output token value given the learned policy of an actor network that has access to expert actions and allows optimization on task specific scores. The action value scores are incorporated into the actor network to improve sequence prediction by using temporal difference methods ~\cite{sutton1988learning}. They also showed that combining AC training with log-likelihood resulted in good performance.

\textit{Reward Augmented Maximum Likelihood} (RML) ~\cite{norouzi2016reward} combine conditional log-likelihood and reward objectives while showing that highest reward is found when the conditional is proportional to the normalized exponentiated rewards. Hence, these scaled rewards are used for smoothing the predicted distribution over states and so the normalized rewards effect the sampling of predicted outputs (tokens in our case). The resulting RML loss is given as $ \sum_{(x,y)\in \mathcal{D}} \{-\sum_{y \in Y}q(y| y^{*}; \tau)\log p_{\theta}(y | x)\}$ where $q(y| y^{*}; \tau)$ is the exponential payoff distribution with temperature parameter $\tau$. This is essentially an expected reward weighted conditional log-likelihood. We too consider the notion of interpolating between the model distribution (i.e scheduled sampling) and reward (i.e dynamically updating the neighborhood distribution via temperature).

\subsection{Sampling Methods in Supervised Learning}
\paragraph{Scheduled Sampling}
The most relevant work to ours is \textit{scheduled sampling} ~\cite{bengio2015scheduled} which also uses an online-based sampling strategy to reduce compounding errors. To alleviate this problem, they propose to alternate between $\hat{y}_{t-1}$ and $y_{t-1}$ using a sampling schedule whereby the probability of using $\hat{y}_{t-1}$ instead of $y_{t-1}$ increases throughout training, allowing the learner to generate multi-step ahead predictions by improving the models robustness with respect to its own prediction errors. 
\newline
A similar method called \textit{Mixed Incremental Cross-Entropy Reinforce} (MIXER) applies REINFORCE for text generation combined with the same idea of using a mixed expert policy (originally inspired by DAgger, but also used in SEARN), whereby incremental learning is used with REINFORCE and cross-entropy (CE). During training, the policy gradually deviates from $\pi^{*}$ provided using CE, to using its own past predictions. The main difference between MIXER and scheduled sampling is that the former relies on past target tokens in prediction and the latter uses REINFORCE to determine if the predictions lead to a sufficiently good return from a task-specific score. In practice however, it is difficult to train a model using MIXER since no reward is given early during training when predictions are near random since there is little to no gradient on the reward.
The above two ideas are interrelated with all aforementioned imitation learning algorithms, which also interpolate between an expert policy and a learned policy. The largest distinction to be made is between how the schedule is chosen for sampling. In this work, we instead dynamically interpolate between the model past predictions, past targets and the additional intermediate step of using past target neighbors and use individual schedules for the latter two. We now detail our methodology. 

\section{Methodology}\label{sec:method}

\subsection{Model Description}
For a given input-output pair $(X, Y)$ of sequence length $T$, where $X$ is parameterized by $\theta$, the log-likelihood is given as \autoref{eq:log_prob} where $\alpha \in [0, 1]$ is a heuristic introduced to control the trade-off between generating longer or shorter sentences ($\alpha < 1$ penalizes longer sentences).

\begin{equation}\label{eq:log_prob}
\frac{1}{T^{\alpha}}\sum_{t = 1}^{T} \log p(y_t|y_{1:t-1}, X; \theta)
\end{equation}

For a given word $x_t \in X$ at timestep $t$ we create an encoded hidden vector $h_t$ parameterized by $\theta$ to generate prediction $\hat{y}_t$. For language models this entails a linear projection to the size of the vocabulary followed by a normalization function such as a $\mathtt{softmax}$ denoted as $g(\cdot)$ to form a probability distribution. The recurrent neural network recursive function we denote as $f_\theta(\cdot)$ herein. 
Treating language modeling as one of classification is common practice where the previous hidden state $h_{t-1}$ of the model and previous output $y_{t-1}$ are used to produce the current hidden state $h_t$ i.e $p(h_t|h_{t-1} y_{t-1};\theta)$. An end of line identifier is used to allow for variable length sequence, often sequences with similar length are grouped together followed by padding, allowing for more efficient mini-batch processing. Stochastic gradient descent is used to maximize the log likelihood of producing the correct tokens in the sequence $Y$.

At test time, target outputs $Y$ are not available, therefore the model uses its current prediction $\hat{y}_{t}$ with hidden state $h_t$ to predict $p(\hat{y}_{t+1}|\hat{y}_{t}, h_{t};\theta)$. As mentioned, this can lead to an accumulation of errors. The current prediction $\hat{y}_{t}$ can be obtained by either acting greedily and selecting the most probable word or sample from the distribution from the softmax output. Alternatively, beam search can be used to consider $b$ number of most likely predictions which means there are $T^{b}$ possible generations for a sequence. By storing an $n$-gram transition lookup table we can check which expansions of the considered beam search terms are most likely throughout the predictions e.g for bigrams we find $p(y_t, y_{t-1}|x)$. In practice, the beam search width $b$ is low (e.g $10 < b < 200$) because the possible number of transitions is $b^V$. 


\subsection{Addressing Compounding Errors}
Teacher forcing refers to when a sequence predictor is given full supervision during training as assumed above. This is the most common way of training neural language models. However, this does not reflect sequence generation at test time since targets are not provided and the model has to rely on its past predictions to generate the next word. 

The aforementioned scheduled sampling method addresses this by alternating between $y_{t-1}$ and $\hat{y}_{t-1}$ at training time so to encourage the model to better predict multiple steps ahead with its own predictions at test time. The tradeoff is controlled with a probability $\epsilon$ such that with probability $(1 - \epsilon_i)^{2}$ the model chooses $\hat{y}_{t-1}$ for the $i^{th}$ mini-batch. $\epsilon$ is then incrementally increased during training as a curriculum learning strategy where the model uses more true targets at the beginning of training and gradually changes to using its own predictions during learning. There are three ways in which $\epsilon$ is set, (1) a linear decay, (2) an exponential decay or (3) an inverse sigmoid decay. 
However, as noted in previous work~\cite{huszar2015not}, scheduled sampling has the limitation that it can lead to learning an incorrect conditional distribution. Additionally, it is assumed that the teacher is optimal when using full supervision. Therefore, we consider the neighbors of words as a $3^{rd}$ alternative that helps the teacher guide the learner. We too test a linear decay, exponential decay and sigmoid decay, for both neighborhood sampling and prediction sampling while introducing some non-parametric alternatives as shown in \autoref{eq:p1}, \ref{eq:p2} and \ref{eq:p3}.

\begin{gather}
	g_1(x) = 1 - x/e^{-x} \label{eq:p1}\\
	g_2(x) = 1/2 + \sin (x\pi - \pi/2) \label{eq:p2}\\
    g_3(x) = 2/(e^{10x}+1) \label{eq:p3}
\end{gather}

During training the $k$-neighbors sampling probability can be updated by shifting mass to neighbors that were replaced during epochs and produced a higher difference in expected reward (i.e validation perplexity) between epochs by tuning for temperature $\tau$ in the exponential payoff shown in \autoref{eq:payoff}. Here, $r(\cdot, \cdot)$ is the reward function that, in our case, calculates the validation perplexity $p$ at each $i^{th}$ epochs. In experimentation we initialize $\tau=0.1$ to begin conservative and allow the model to explore further neighbors as the model learns. Updates are carried out after each epoch for all samples in the training data.

\begin{equation}\label{eq:payoff}
\phi(y,\hat{y},\tau) = \frac{\exp(r(y, \hat{y})/\tau)}{\sum_{y \in Y}\exp(r(y, \hat{y})/\tau)}
\end{equation}

This can be considered as an auxiliary task optimized for expected reward. However, this does not effect the monotonic sampling rate for choosing elements within a mini-batch, only the choice of replacement in NNRS. In \autoref{alg:nnrs} we see that a simple difference update rule is used where if the current validation perplexity $p_i$ is better than $p_{i-1}$ the temperature is increased. This means as the model begins to improve validation perplexity, the model begins to explore neighbors that are further in cosine distance to the target word $y_t$.

\subsubsection{Sampling Strategies}

The probability of token $y$ in a sequence $Y$ being chosen for replacement is directly proportional to the epoch number (i.e higher sampling rate near the end of training), denoted as subscript $i$ in \autoref{alg:ssa}. For all unique words $v \in V$ the cosine similarity function $m(\cdot)$ is used to compute the cosine similarity between word vectors to obtain the top $k$ neighbors $\forall v \in V$, as shown in \autoref{alg:nnrs} where $\phi = e^{z}/\sum_{k} e^{z_{k}}$ and $d_\mathcal{N}$ is a sorted list of probabilities. In practice it is not necessary to loop $\forall v \in V$, it is instead stored in a lookup table with the corresponding sampling probability and word index for replacement.

\begin{figure}[ht]
\begin{center}
 \includegraphics[scale=0.5]{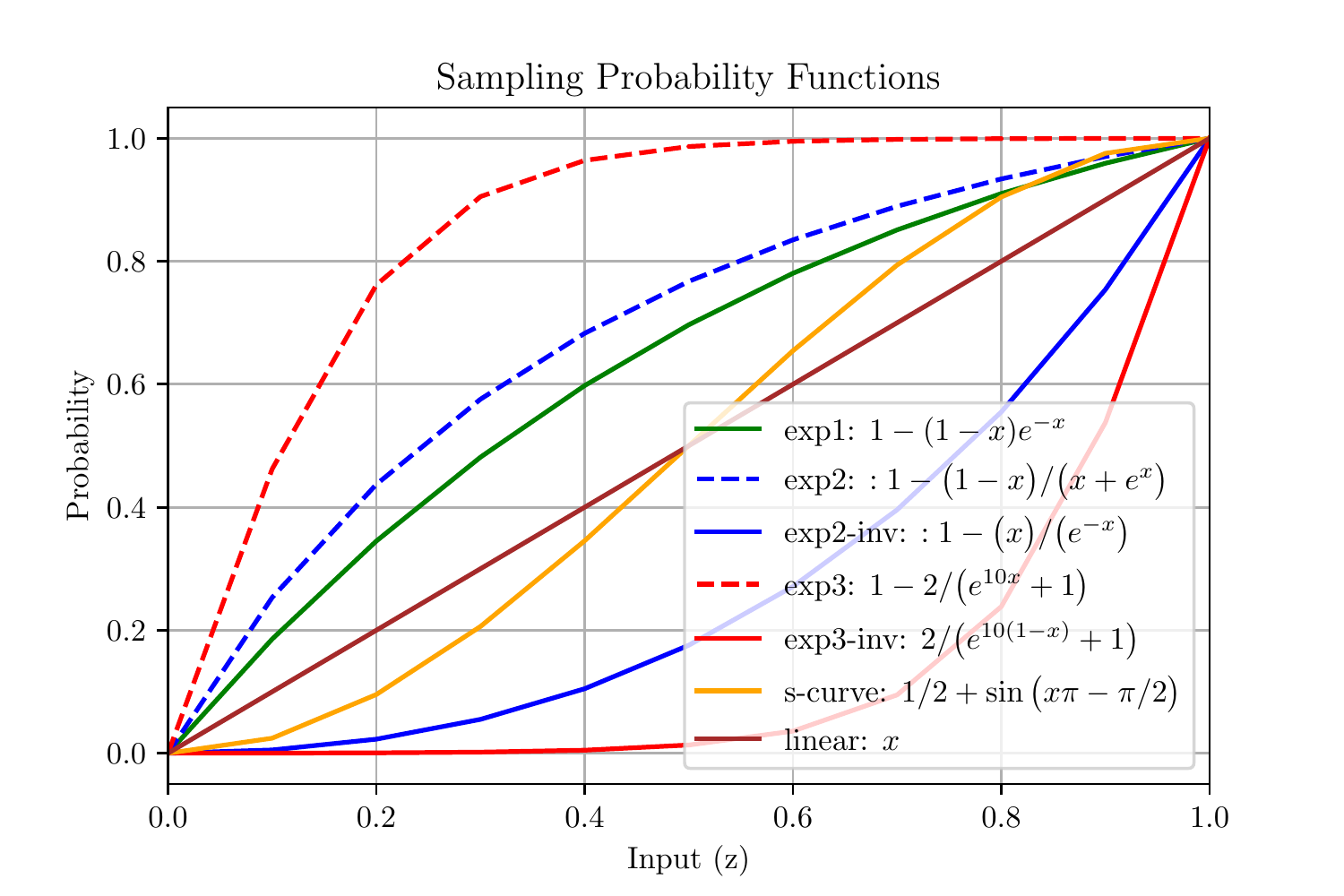}
 \caption{Functions For Sampling Probabilities ($x$ represents the 0-1 normalized number of epochs)}\label{fig:spf}
\end{center}
\end{figure}

\begin{algorithm}[ht]
    \SetKwInOut{Input}{Input}
    \SetKwInOut{Output}{Output}
    \underline{NNRS} $(x_t, k, m, p, \tau)$\;
    \Input{$y_t$, $k$ neighbors, sim. measure $m$, val.perplexities $p$, \iffalse \epoch $i$,\fi temperature $\tau$} 
    \Output{$\tilde{Y}$ Neighbor Replacement Target}    
		\For{$v \in \mathcal{V}$}{
			$d_{\mathcal{N}} \leftarrow m(v, x_t)$\;
    	}
    $\mathcal{N}_{y_t} \leftarrow  \phi(d_\mathcal{N}[:k], \tau, p_i)$\;
    $\tilde{y}_t \sim \mathcal{N}_{y_t}$\;
    \If{$(p_i>p_{i-1}) \; \&\& \; (\tau < 1)$}
     {
       $\tau \leftarrow \tau + |\tau - (2^{\tau}-1)|$\;
     }
     \Else{
      $\tau \leftarrow 2^{\tau}-1$\;
     }
    return $(\tilde{x}_t, \tau)$\;	    
    \caption{N-Neighborhood Replacement Sampling}
    \label{alg:nnrs}
\end{algorithm}

Samples are then drawn from this truncated probability distribution of the $k$ nearest neighbors using cosine similarity function $m$ between word vectors of the targets $Y$ of sequence $X$, as shown in \autoref{alg:ssa}. There are two methods to consider here. 

The $k$ is chosen such that $k \approx \log_2(|\mathcal{V}|)$ which is sample efficient and speeds up sampling at run time. This strategy can be interpreted as the \textit{teacher} coaching the \textit{learner} when the learner has difficulty predicting from the teacher policy. The assumption is that synonymous words provide alternative paths that can lead to better expected perplexity.

\begin{algorithm}
    \SetKwInOut{Input}{Input}
    \SetKwInOut{Output}{Output}
    \underline{SSA} $(X,Y, n, g_{ss}, g_{nnrs})$\;
    \Input{Input sequence $X$, target sequence $Y$, training fraction $n$, sampling functions $g_{ss},g_{nnrs}$}
    \Output{$\hat{Y}_{y \sim \mathbb{U}}$ Neighbor Replacement Target}
    
    $\epsilon_i \leftarrow g_{ss}(n) \;, \gamma_i \leftarrow g_{nnrs}(n)$\; 
    $\tau = 0.1, \; k = \log_{2}(|V|), \; m = \cos(.,.)$\; 
   	\For{$t \in T$}{
    	$[\pi_1, \pi_2]$ $\sim \mathbb{U}$ 
      
      \If(\tcp*[f]{$Choose \; randomly$}){
      	$(\epsilon_i>\pi_1) \;\&\; (\gamma_i>\pi_2)$
      }
      {
        $x_{t} \xleftarrow[]{{\mathbb{U}}} [\hat{y}_{t-1}, \tilde{y}_{t-1}] $\;
      }

      \ElseIf(\tcp*[f]{$Assign\; past\; prediction$}){
      	$\epsilon_i>\pi_1$
        }
      {
        $x_{t} \leftarrow \hat{y}_{t-1}$\;
      }
      \ElseIf(\tcp*[f]{$Assign \; sampled\; target\; neighbor$}){$\gamma_i>\pi_2$}
      {
      	$\tilde{y}_{t-1}, \tau \leftarrow \mathtt{NNRS}(y_{t-1}, k, m, p, \tau)$\;
        $x_{t} \leftarrow \tilde{y}_{t-1}$
      }
      \Else{
       $x_{t} \leftarrow y_{t-1}$
      }
      $\hat{y}_{t}, h_{t} \leftarrow f_{\theta}(x_{t}, h_{t-1})$\;
  	  $\hat{Y} \cup \hat{y}_{t}$\;
    }
    
    return $(\hat{Y})$\;
	
    \caption{Sampling Strategy Algorithm}
    \label{alg:ssa}
\end{algorithm}

The main motivation for carrying out NNRS is to address the aforementioned problem of compounding errors. In such cases, we would like the model to follow a stochastic policy $\tilde{\pi}$ instead of the original teacher policy $\pi^{*}$. This can also be considered as an exploration strategy that increases over batches as the teacher (or critic in actor-critic networks) teaches the learner. Additionally we consider updating the $k$ truncated probability distribution matrix based on the validation performance during training. Concretely if the synthetic targets $\tilde{Y}$ are found to improve validation perplexity score we update the probability distribution to give more mass to the sampled neighbors. The method makes the assumption that because the model has relatively less data to learn from, providing local neighbors for input replacement can help the model learn from \say{easier to predict} samples (i.e curriculum learning) before using its own predictions as is the case for scheduled sampling. 

\begin{figure}
\centering
\captionsetup{justification=centering}
 \includegraphics[scale=1.0]{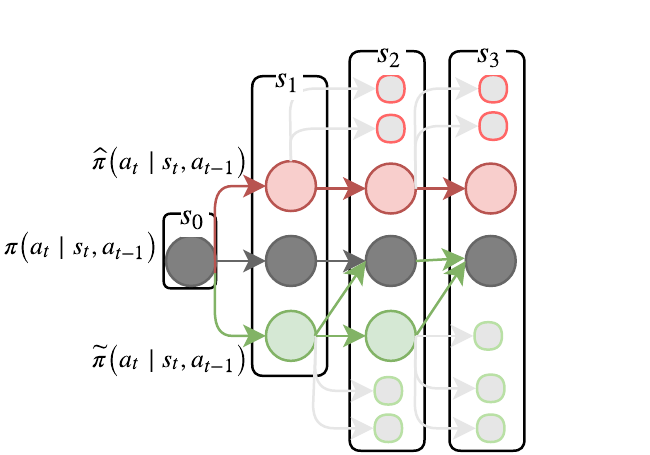}
 \caption{Interpolating between teacher (black), predictions (red) and teacher's neighbors (green) for target sequence $Y$}\label{fig:stochastic_policy}
\end{figure}

\autoref{fig:stochastic_policy} illustrates the trajectories the sequence might take under a given policy from an initial state $s_0$. We show an example where past predictions can lead to a divergence from the teacher trajectory. This can be due to a \textit{difficult to learn} transition from $s_0 \to s_1$, but given a synonymous neighbor (green) as the local one-step deviation allows us to better guide the learner, thus making the transition from $s_1 \to s_2$ easier. Also, we would expect the output to be less sensitive to perturbations in the input since the input is locally bounded by the space occupied by the $k$ nearest neighbors of the target $\tilde{y}_{t}$. Likewise, $\tilde{y}_{t}$ can be considered as emulating the problem of compounding errors, since the conditional probability $p(y_{t+1}|x_{1:t},\tilde{y}_{t};\theta)$ is conditioning on the sampled neighbor $\tilde{y}_{t}$ instead of the true target $y_{t}$.

In cases where the model finds it difficult to transition from using $y_{t-1} \to \hat{y}_{t-1}$, interpolating with the neighborhood samples $\tilde{y}$ can provide a smoother policy ($y \to \tilde{y} \to \hat{y}$). Hence, the method can be considered as a smoothing method which assigns some mass to unseen transitions, similar to the purpose of Laplacian smoothing except, except it is bounded by $k$ neighbors which is natural and directly proportional to transition probabilities. 
In the case of a static sampling rate $[\epsilon, \gamma]$ we can define the log-likelihood loss with scheduled sampling and NNRS in terms of KL-divergences, as shown in \autoref{eq:ss_nnrs}. $P_{x_{t}}$ and $Q_{x_{t}}$ are the marginal distributions for input token $x_{t}$ while $P_{x_{t}|x_{t-1}=h}$ and $Q_{x_{t}|x_{t-1}=h}$ represent the conditional distributions. 

\begin{equation}\label{eq:ss_nnrs}
\begin{gathered}
	D_{SS-NNRS}[P||Q] = KL[P_{x_{t-1}}||Q_{x_{t-1}}] + \underbrace{(1 - \epsilon)\mathbb{E}_{h \sim Q_{x_{t-1}}}}\\
    \underbrace{KL[P_{x_{t-1}}||Q_{x_{t}|x_{t-1}=h}] + \epsilon \mathbb{E}_{h \sim P_{x_{t-1}}}KL[P_{x_{t}|x_{t-1}}|| Q_{x_{t}|x_{t-1}}] }_\text{scheduled sampling}\\
    + \underbrace{(1 - \gamma) \mathbb{E}_{h \sim Q_{x_{t-1}}} KL[P_{x_{t+1}}||Q_{x_{t}|x_{t-1}=h}]}\\ 
    +  \underbrace{\gamma \mathbb{E}_{h \sim P_{x_{t-1}}} KL[P_{x_{t}|x_{t-1}}|| Q_{x_{t}|x_{t-1}}]}_\text{nearest replacement sampling}
 \end{gathered}
\end{equation}

For optimization, Stochastic Gradient Descent with cosine annealing~\cite{reddi2018convergence} of the learning rate is used, as shown in \autoref{eq:cos_annealing} where $\alpha_0$ is the initial learning rate and $\mathcal{T}_{sub}$ is the period between restarts, in our case $\mathcal{T}_{sub}=\mathcal{T}_{max}$.

\begin{equation}\label{eq:cos_annealing}
\alpha_t = \alpha_{\min} + \frac{1}{2}\Big((\alpha_{0} - \alpha_{\min})
(1 + \cos(\pi \frac{\mathcal{T}_{sub}}{\mathcal{T}_{\max}})\Big)
\end{equation}

\subsubsection{Connections To Neighborhood Sampling}
Imitation learning algorithms such as DAgger grow the dataset size as new trajectories are created for each policy. Our proposed NNRS method is easily applied at training time and only requires a fast neighborhood sampling step online for each batch. NNRS does not sample from potential neighbors that are outside the vocabulary, only the nearest neighbors within the vocabulary (to avoid snooping). In comparison to LOLS, we do not require a separate actor network to guide the learner (i.e actor) but rather rely on a curriculum learning based stochastic policy where the expected return is given by the validation perplexity score which is accounted for by adjusting temperature $\tau$ throughout training. As an exploration strategy, $\tau$ is increased throughout training so that neighbors that are further away are gradually assigned a higher sampling probability. LOLS optimizes on the regret to the teacher policy and its own one step deviations. In contrast, this approach optimizes between an interpolation of the regret to the teacher policy, regret to teachers one step deviations and regret to teacher policy with multi-step ahead predictions. When NNRS is used with SS it results in the model being robust to its prediction deviations from the original input but also the locally sampled neighborhood of the input.

\section{Experiments}

\begin{table*}
\centering

\resizebox{1.0\linewidth}{!}{%

\begin{tabular}{c|cccc?cc|cc|cc|cc?cc|cc|cc|cc}
 \toprule

Configuration  & \multicolumn{4}{c|}{Parameter Setting} & \multicolumn{8}{c|}{Wiki-102} & \multicolumn{8}{c}{Penn-Treebank}\\

\midrule

\multicolumn{5}{c|}{} & \multicolumn{2}{c|}{Linear} & \multicolumn{2}{c}{S-Shaped Curve} & \multicolumn{2}{|c}{Exponential Increase}& \multicolumn{2}{|c|}{Static}  & \multicolumn{2}{c|}{Linear} & \multicolumn{2}{c}{S-Shaped Curve} & \multicolumn{2}{|c}{Exponential Increase}& \multicolumn{2}{|c}{Static}\\

\midrule

& $\epsilon_s$ & $\epsilon_e$ & $\gamma_s$ & $\gamma_e$ & Valid & Test & Valid & Test & Valid & Test & Valid & Test & Valid & Test & Valid & Test & Valid & Test & Valid & Test\\
\midrule 


No Sampling & - & - & - & - & 140.28 & 128.78 & 140.28 & 128.78 & 140.28 & 128.78 & 140.28 & 128.78 & 76.25 & 71.81 & 76.25 & 71.81 & 76.25 & 71.81 & 76.25 & 71.81 \\
\midrule
TPRS-1 & 0 & 0 & 0 & 0.2 & \cellcolor{black!20}143.78 & \cellcolor{black!20}131.18 & \cellcolor{black!20}137.69 & \cellcolor{black!20}127.05 & 137.63 & 136.88 & 136.31 & 126.49 & \cellcolor{black!20}81.42 & \cellcolor{black!20}76.58 & \cellcolor{black!20}83.40 & \cellcolor{black!20}81.22 & 77.56 & 73.02 & \cellcolor{black!20}81.45 & \cellcolor{black!20}80.36\\
TPRS-2 & 0 & 0 & 0 & 0.3 & 154.93 & 144.02 & 146.92 & 137.07 & \cellcolor{black!20}137.11 & \cellcolor{black!20}125.41 & \cellcolor{black!20}137.10 & \cellcolor{black!20}125.95 & 96.91 & 94.30 & 88.79 & 86.17 & \cellcolor{black!20}77.63 & \cellcolor{black!20}72.80 & 91.73 & 84.48\\
TPRS-3 & 0 & 0 & 0 & 0.5 & 159.11 & 148.41 & 148.10 & 139.58 & 138.46 & 127.97 & 138.53 & 129.87 & 97.62 & 94.78 & 88.46 & 86.05 & 76.60 & 72.77 & 98.31 & 95.23\\


\midrule

NNRS-1 & 0 & 0 & 0 & 0.2 & \cellcolor{black!20}142.53 & \cellcolor{black!20}130.45 & \cellcolor{black!20}137.08 & \cellcolor{black!20}126.82 & 136.81 & 136.11 & 135.06 & 136.02 & \cellcolor{black!20}80.91 & \cellcolor{black!20}76.70 & \cellcolor{black!20}83.17 & \cellcolor{black!20}79.62 & \cellcolor{black!20}76.00 & \cellcolor{black!20}72.19 & \cellcolor{black!20}82.23 & \cellcolor{black!20}80.03\\
NNRS-2 & 0 & 0 & 0 & 0.3 & 154.50 & 143.13 & 149.69 & 136.98 & \cellcolor{black!20}136.83 & \cellcolor{black!20}125.53 & 136.34 & 125.84 & 96.66 & 93.18 & 88.68 & 85.29 & 77.12 & 73.05 & 92.12 & 84.29\\
NNRS-3 & 0 & 0 & 0 & 0.5 & 158.30 & 147.13 & 148.02 & 137.34 & 137.56 & 127.61 & 132.62 & 121.50 & 96.95 & 93.15 & 88.21 & 84.47 & 75.91 & 72.46 & 97.55 & 94.79\\

\midrule
SS-1 & 0 & 0.2 & 0 & 0 & 139.31 & 128.50 & 143.82 & 131.08 & 137.72 & 126.46 & 135.55 & 122.61 & 83.63 & 80.03 & \cellcolor{black!20}82.33 & \cellcolor{black!20}79.02 & 76.71 & 73.32 & 74.50 & 70.20\\
SS-2 & 0 & 0.3 & 0 & 0 & 137.78 & 126.00 & 138.39 & 126.80 & \cellcolor{black!20}135.89 & \cellcolor{black!20}125.03 & 131.29 & 121.52 & \cellcolor{black!20}94.74 & \cellcolor{black!20}79.82 & 84.42 & 80.03 & 76.43 & 73.46 & \cellcolor{black!20}74.56 & \cellcolor{black!20}70.25\\
SS-3 & 0 & 0.5 & 0 & 0 & \cellcolor{black!20}135.14 & \cellcolor{black!20}124.29 & \cellcolor{black!20}136.88 & \cellcolor{black!20}125.41 & 136.98 & 125.96 & \cellcolor{black!20}131.28 & \cellcolor{black!20}121.51  & 92.48 & 88.92 & 85.87 & 82.37 & 76.41 & 73.15 & 74.11 & 69.48\\
SS-4 & 0 & 0.8 & 0 & 0 & 140.97 & 130.00 & 141.13 & 129.99 & 138.09 & 127.23 & 134.32 & 122.86 & 92.94 & 88.85 & 87.29 & 84.27 & \cellcolor{black!20}76.22 & \cellcolor{black!20}72.98 & 74.02 & 70.23\\

\midrule
SS-NNRS-1 & 0 & 0.2 & 0 & 0.2 & 139.32 & 128.50 &  \cellcolor{black!20}141.57 & \cellcolor{black!20}129.67 & 137.73 & 126.47 & 135.55 & 122.60 & \cellcolor{black!20}83.62 & \cellcolor{black!20}80.03 & \cellcolor{black!20}81.99 & \cellcolor{black!20}77.86 & 76.71 & 73.31 & 74.50 & 70.20\\
SS-NNRS-2 & 0 & 0.3 & 0 & 0.3 & 137.78 & 126.01 & 147.34 &135.61 & 136.02 & 125.73  & 137.73 & 126.47 & 95.39 & 92.18  & 87.97 & 84.64 & 75.43 & 72.46 & 74.64 & 70.43\\

SS-NNRS-3 & 0 & 0.5 & 0 & 0.2 & \cellcolor{black!20}135.14 & \cellcolor{black!20}124.29 & 149.00 & 137.97 & \cellcolor{black!20}135.82 & \cellcolor{black!20}124.72 & \cellcolor{red!20}130.95 & \cellcolor{red!20}120.76 & 95.90 & 92.96 & 88.56 & 84.96 & \cellcolor{black!20}74.83 & \cellcolor{black!20}70.95 & \cellcolor{red!20}72.89 & \cellcolor{red!20}69.06\\

SS-NNRS-4 & 0.2 & 0.5 & 0.2 & 0.5 & 150.97 & 138.59 & 146.01 & 134.67 &  135.88 & 126.02 & 123.84 & 121.98 & 96.78 & 93.20 & 88.56 & 87.06 & 76.41 & 73.15 & 74.42 & 69.88\\

SS-NNRS-5 & 0 & 0.5 & 0 & 0.5 & 136.22 & 125.70 & 151.39 & 138.44 & 137.02 & 125.84 & 132.17 & 121.86 & 97.12 & 93.90 & 90.37 & 86.25 & 76.02 & 72.31 & 74.08 & 70.79\\

SS-NNRS-6 & 0 & 0.8 & 0 & 0.2 & 148.96 & 130.01 & 141.14 & 129.99 & 138.09 & 127.24 & 134.32 & 122.89 & 95.91 & 92.87 & 81.89 & 78.54 & 74.74 & 70.64& 74.02 & 70.23 \\

SS-NNRS-7 & 0.2 & 0.8 & 0.2 & 0.5 & 155.17 & 148.21 & 149.58 & 137.83 & 135.45 & 124.82 & 131.09 & 121.43 & 96.77 & 93.64 & 88.27 & 85.04 & 76.35 & 72.74 & 74.43 & 70.12\\

\bottomrule
\end{tabular}%
}
\vspace{1em}
\captionsetup{justification=centering}
   \caption{Perplexity for Transition Probability Sampling (TPRS), Scheduled Sampling (SS) and Nearest-Neighbor Replacement Sampling (NNRS) with linear, s-shaped curve and exponential sampling functions for a standard 2-hidden layer LSTM}
  \label{tab:nsr_results}
\end{table*}

\subsection{Training Details}
We use a standard 2-hidden layer LSTM sequence model ~\cite{sundermeyer2012lstm} as the basis of our analysis and purposefully avoid using any additional mechanisms (e.g attention) because we aim at directly evaluating the effect of sampling strategies to NLM.We test different configurations for sampling strategies, namely: an LSTM with scheduled sampling with varying $\epsilon$, LSTM with NNRS with varying $\gamma$ and an LSTM with both NNRS and scheduled sampling. The main baseline is the same LSTM network with no sampling and an LSTM that uses NNRS where the top $k$ neighbors are chosen directly from the transition probability matrix (denoted as TPRS). The former is the main baseline, as the focus of the paper is to identify such sampling techniques improve over the standard model with identical hyperparameter settings. The latter baseline further establishes if creating top $k$ neighbors based on cosine similarity of embeddings is a good approximate to replacement based on the full transition probability matrix. 
The LSTM consists of 2-hidden layers with a embedding dimension and hidden dimension size $|x|=|h|=200$. Xavier uniform initialization is used with $(\mu = 0, \sigma = 0.1)$ and gradients are clipped at $0.5$ threshold if exceeded at each update, with a batch size $|X_s|=30$.

Encoder and decoder weights are tied in the network which reduces the number of parameters and therefore can be considered a form of regularization. From an information theoretic point of view, ~\cite{shwartz2017opening} showed that the encoder performs input data compression with high mutual information between the input and encoder for the most epochs while the decoder has lower mutual information with the target. This is then reversed towards the end of training whereby the information in the network diffuses. We argue this what allows us to tie weights and still perform relatively well since most time is taken efficiently representing the input and less time spent on fitting the training labels, particularly relevant in NLM's since the input and output share the same space. 

For scheduled sampling and NNRS we consider a range of functions that represent the curriculum learning behavior during training as illustrated in \autoref{fig:spf}. This Figure shows the cumulative distribution for each function, however this assumes a starting NNRS and scheduled sampling probability $[\epsilon_s, \gamma_s]=0$ and ending probability $[\epsilon_e, \gamma_e] = 1$ respectively. In experimentation, we report various start and end ranges for both $\epsilon$ and $\gamma$ with these functions.

\begin{figure}
\centering
\captionsetup{justification=centering}
 \includegraphics[scale=0.5]{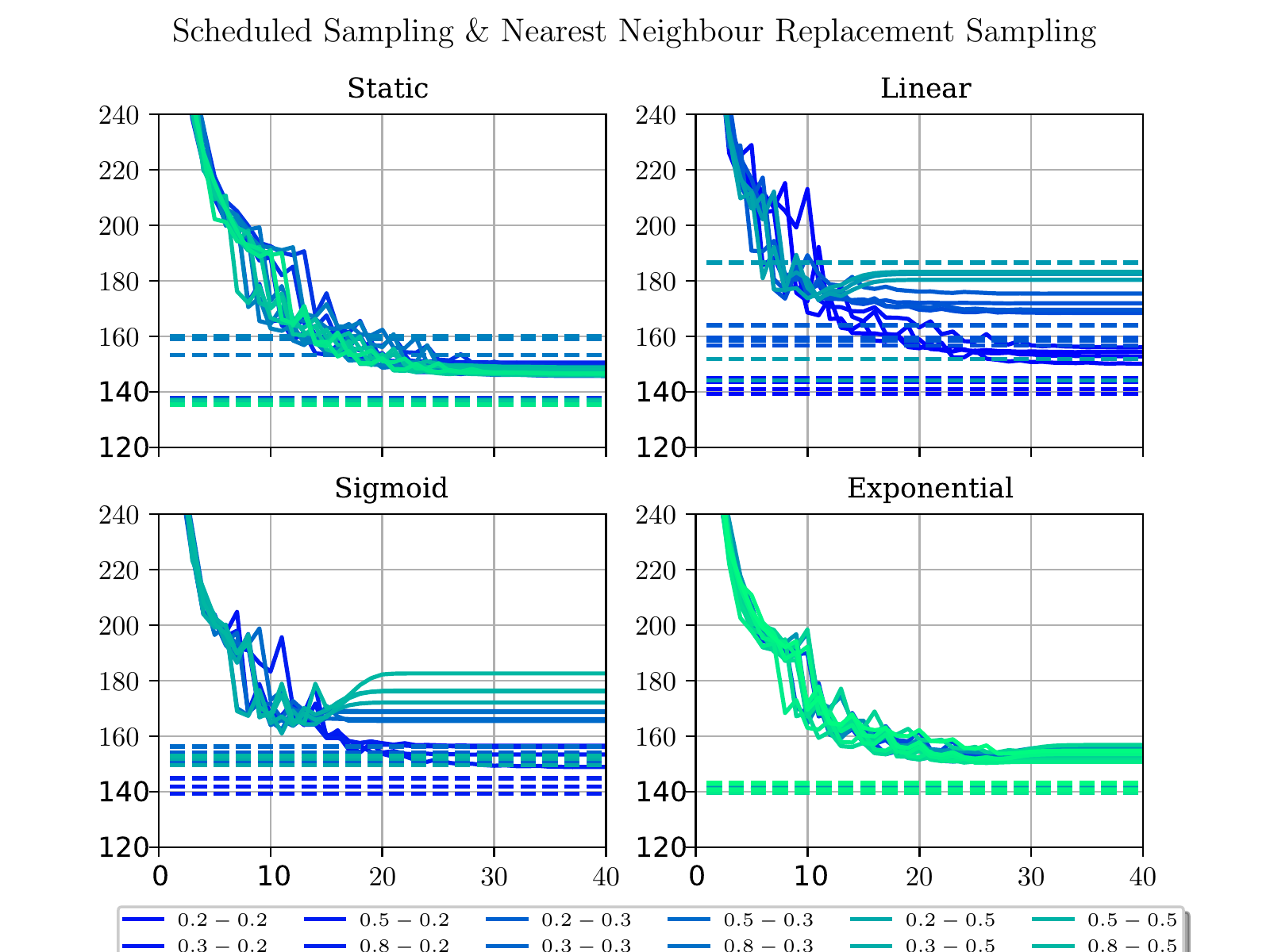}
 \caption{Perplexity on WikiText2: Scheduled Sampling and NNRS (best viewed in color)}\label{fig:wiki2_lc_ss}
\end{figure}

\subsection{Experimental Results}
\paragraph{WikiText-2 Performance}
\autoref{fig:wiki2_lc_ss} shows the perplexity scores for the different parameter settings for $\epsilon$ and $\gamma$ over 40 training epochs. Solid lines indicate validation perplexity scores throughout training and dashed horizontal lines indicate corresponding test perplexity score. We find that most learning is carried out after 20 epochs given that the learning rate is initially high ($\alpha=20$) and annealed according to \autoref{eq:cos_annealing}. The optimal settings are found with $\epsilon = 0.5$ and $\gamma = 0.2$ with a static probability sampling rate. Additionally, using an exponential sampling rate (denoted as exp3-inv in \autoref{fig:spf}) outperforms the linear and sigmoid functions. It would appear that this is because the exponential function carries out the majority of sampling late in training when the model has reached learning capacity from the teacher policy. As shown, with linear and sigmoid functions, when the sampling rate is high too early, the resultant learning policy is too stochastic (between 10-15 epochs) and the learning rate is annealed early on. In contrast, since the exponential function carries out most sampling later in training, it allows for a higher upper bound on $\gamma$ and $\epsilon$, as seen in light green.

\begin{figure}
\centering
\captionsetup{justification=centering}
 \includegraphics[scale=0.5]{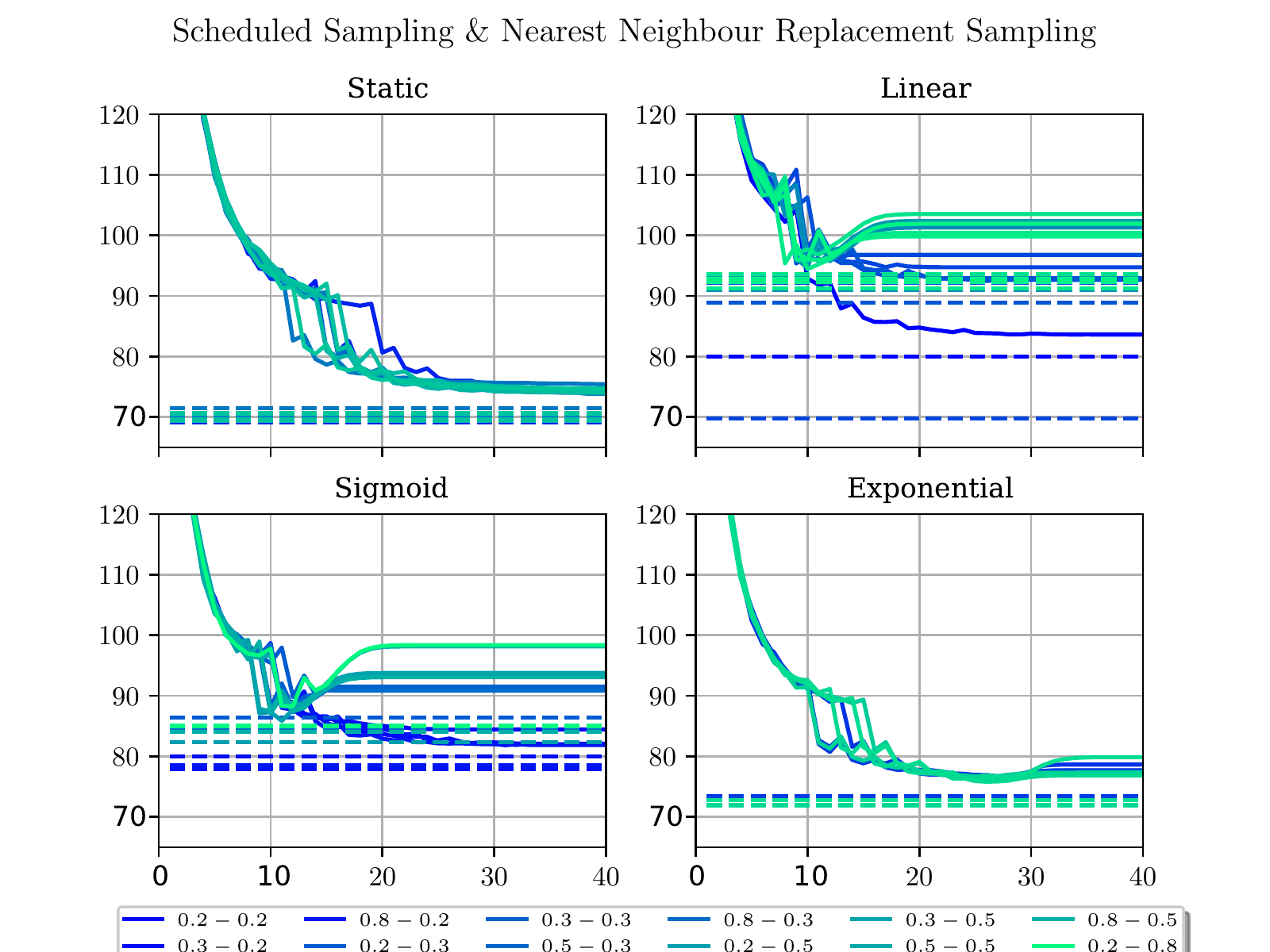}
 \caption{Perplexity on Penn-Treebank: Scheduled Sampling and NNRS}\label{fig:ptb_lc_ss}
\end{figure}

\paragraph{Penn-Treebank Performance}
\autoref{fig:ptb_lc_ss} shows the results of the proposed approach on Penn-Treebank. We find that static and exponential settings for both SS and NNRS show best performance and converge quicker. In all cases, convergence for all functions behaves as expected for $[\sigma, \gamma] < 0.3$. We also see the same trend as with WikiText-2, that exponential function allows for higher sampling rates when compared to linear and sigmoid functions. Again, this suggests that NNRS is most effective near convergence, as the sampling probability exponentially increases while the validation perplexity scores begin to plateau over epochs.  

\paragraph{Schedule Parameter Grid Search Results}
\autoref{tab:nsr_results} shows the results of the model with varying $\epsilon$ and $\gamma$ upper and lower thresholds using an linear, s-curve, exponential and static sampling strategy for all tested datasets. Overall, we find that using scheduled sampling in conjunction with neighborhood replacement sampling yields best performance for both datasets. Best performance for both Wikitext-2 and PTB is found with $\epsilon = (0, 0.5)$ and $\gamma (0, 0.2)$, and slightly improves over only using scheduled sampling.  For PTB, $\gamma_e = 0.2$ performs the best for linear and sigmoid functions, $\gamma = 0.5$ for exponential and static sampling rates and overall a constant sampling rate. 

Although, the normalized exponential distribution better than the sigmoid and linear scheduled for both $\epsilon$ and $\gamma$ on both datasets. It would seem that this is because the majority of neighbor replacements are towards the end of training when the learner has already reached capacity in what can be learned from the teacher policy. Moreover, this would approximately be the inverse of the normalized validation perplexity. This also coincides with the theoretical guarantees of using an exponential decay schedule provided in DAgger ~\cite{ross2010efficient}. Overall, interpolating between neighborhood replacement sampling and scheduled sampling produces the best performance on average for both datasets. A low constant sampling rate has yielded best performance for both datasets. In particular, we find that controlling the temperature $\tau$ based on the validation perplexity allows for exploration to allow all $k$-neighbors to be visited. 
When compared to using no sampling there is an 8 point perplexity decrease using our approach on WikiText-2 and a 2.75 test perplexity decrease on Penn-Treebank. In comparison to using the full transition probability for replacement (i.e TPRS) we see slight improvements in using proposed approach while requiring less memory to store readily available pretrained embeddings.

\section{Conclusion}

We presented a curriculum learning sampling strategy to mitigate compounding errors in sequence-to-sequence models. We find performance improvements particularly for sampling functions that generate monotonically increasing sampling rates that are inversely proportional to the slope of decreasing validation performance.
This is empirically demonstrated when comparing a standard 2-hidden layer LSTM (with identical hyperparameter settings) with the following settings: (1) no sampling strategies, (2) a baseline transition probability replacement sampling, (3) the proposed nearest neighbor replacement sampling technique, (4) scheduled sampling and (5) a combination of (3) and (4).

We find the best sampling probability settings for each dataset along with the corresponding cumulative probability functions and conclude that in general, an exponential increase and static sampling probability during training performs better over sigmoid and linear functions. In other words, a schedule that samples with relatively high probability early in training leads to a performance degradation whereas sampling with high probability towards the end of training can improve performance. 

We also find that test set perplexity scores increase for both datasets when using nearest neighbor replacement sampling in conjunction with scheduled sampling.

\bibliography{aaai}
\bibliographystyle{aaai}

\end{document}